\documentclass[sn-nature]{sn-jnl}

\usepackage[T1]{fontenc}

\usepackage[sfdefault,scale=.95]{atkinson}   
\usepackage[utf8]{inputenc}

\usepackage{graphicx}%
\usepackage{multirow}%
\usepackage{amsmath,amssymb,amsfonts}%
\usepackage{amsthm}%
\usepackage{mathrsfs}%
\usepackage[title]{appendix}%
\usepackage{xcolor}%
\usepackage{textcomp}%
\usepackage{manyfoot}%
\usepackage{booktabs}%
\usepackage{algorithm}%
\usepackage{algorithmicx}%
\usepackage{algpseudocode}%
\usepackage{listings}%
\usepackage{natbib}

\theoremstyle{thmstyleone}%
\theoremstyle{thmstyletwo}%

\theoremstyle{thmstylethree}%

\begin{document}

\title[Article Title]{Who Does What in Deep Learning? Multidimensional Game-Theoretic Attribution of Function of Neural Units}


\author*[1,2]{\fnm{Shrey} \sur{Dixit}}\email{dixit@cbs.mpg.de}
\equalcont{These authors contributed equally to this work.}

\author[1, 3]{\fnm{Kayson} \sur{Fakhar}}
\equalcont{These authors contributed equally to this work.}

\author[1]{\fnm{Fatemeh} \sur{Hadaeghi}}
\author[4]{\fnm{Patrick} \sur{Mineault}}
\author[5,6]{\fnm{Konrad P.} \sur{Kording}}
\author[1,7]{\fnm{Claus C.} \sur{Hilgetag}}

\affil[1]{\orgdiv{Institute of Computational Neuroscience}, \orgname{University Medical Center Eppendorf}, \orgaddress{\city{Hamburg}, \country{Germany}}}

\affil[2]{\orgdiv{International Max Planck Research School on Cognitive Neuroimaging}, \orgaddress{\city{Leipzig}, \country{Germany}}}

\affil[3]{\orgdiv{MRC Cognition and Brain Sciences Unit}, \orgname{University of Cambridge}, \orgaddress{\city{Cambridge}, \country{United Kingdom}}}

\affil[4]{\orgname{Mila - Quebec Artificial Intelligence Institute}, \orgaddress{\city{Montreal}, \country{Canada}}}

\affil[5]{\orgdiv{Learning in Machines \& Brains}, \orgname{CIFAR}, \orgaddress{\city{Toronto}, \country{Canada}}}

\affil[6]{\orgdiv{Departments of Bioengineering and Neuroscience}, \orgname{University of Pennsylvania}, \orgaddress{\city{Philadelphia}, \country{USA}}}

\affil[7]{\orgdiv{Department of Health Sciences}, \orgname{Boston University}, \orgaddress{\city{Boston}, \country{USA}}}

\abstract{
Neural networks now generate text, images, and speech with billions of parameters, producing a need to know how each neural unit contributes to these high-dimensional outputs. Existing explainable-AI methods, such as SHAP, attribute importance to inputs, but cannot quantify the contributions of neural units across thousands of output pixels, tokens, or logits. 
Here we close that gap with Multiperturbation Shapley-value Analysis (MSA), a model-agnostic game-theoretic framework. By systematically lesioning combinations of units, MSA yields Shapley Modes, unit-wise contribution maps that share the exact dimensionality of the model's output.
We apply MSA across scales, from multi-layer perceptrons to the 56-billion-parameter Mixtral-8×7B and Generative Adversarial Networks (GAN). The approach demonstrates how regularisation concentrates computation in a few hubs, exposes language-specific experts inside the LLM, and reveals an inverted pixel-generation hierarchy in GANs.
Together, these results showcase MSA as a powerful approach for interpreting, editing, and compressing deep neural networks.
}

\maketitle

\section{Introduction}\label{sec1}
Over the past decade, neural networks—particularly Large Language Models (LLMs)—have made remarkable strides, with the introduction of ChatGPT marking a turning point in their widespread adoption \cite{wu_brief_2023}. Modern LLMs, such as  GPT-4, estimated to have trillions of parameters, can now process text, speech, and images, reflecting their growing complexity. This progress has outpaced our understanding of their internal workings, raising concerns about transparency and trust. As these systems enter sensitive domains, such as healthcare and finance, interpretability has become increasingly urgent \cite{tjoa2020survey, rudin2019stop}.

The field of explainable AI (XAI) aims to address this challenge, often by attributing model decisions to input features. Among the most popular tools is the SHAP package \cite{lundberg_unified_2017}, which uses game theory \cite{shapley1953value} to quantify feature importance. However, most XAI methods remain focused on inputs, overlooking the contributions of internal neural units. Some recent work explores neuron-level effects on accuracy \cite{ghorbani_neuron_2020}, but lacks generality across architectures or output types, limiting our ability to understand how neural units influence complex outputs like images, audio, or natural language.

For example, in a Stable Diffusion model \cite{rombach_high-resolution_2022}, we might wish to know how each neural unit influences each output pixel. Similarly, understanding how individual inputs influence predictions over time could improve interpretability in finance. And in Natural Language Processing (NLP), tracing unit-level contributions to syntax, semantics, or style could deepen our understanding of language generation.

To address this challenge, we present a general-purpose implementation of Multi-perturbation Shapley-value Analysis (MSA) \cite{keinan_fair_2004}. Our open-source Python package estimates contributions of neural units—whether neurons in multilayer perceptron (MLP) \cite{rumelhart_learning_1986}, filters in convolutional neural network (CNN) \cite{lecun_backpropagation_1989}, or experts in transformer-based Mixture of Experts models \cite{fedus2022switch}—to both scalar and high-dimensional outputs.

MSA works by systematically perturbing neural units, removing or deactivating them, and measuring the effect on the model's output. Each unit is treated as a "player" in a cooperative game, with the output representing the game's "value." The Shapley Value for a unit reflects the unit's average contribution across all possible combinations of intact and perturbed units. For models with multidimensional outputs, such as  images from a deep convolutional generative adversarial network (DCGAN) \cite{goodfellow_generative_2014, radford_unsupervised_2016}, we introduce Shapley Modes, which extend Shapley values to capture unit contributions to individual output elements, for instance, image pixels or language tokens. MSA is also model-agnostic and supports local explanations (specific outputs) and global explanations (aggregated contributions on the network's performance). Whereas Layerwise Relevance Propagation \cite{montavon2019layer}, Integrated Gradients\cite{sundararajan2017axiomatic}, or SHAP provide single-output attributions and struggle with interactions, MSA efficiently quantifies causal unit interactions across high-dimensional outputs.

We demonstrate the utility of MSA through three use cases. We begin by exploring the most basic unit of a neural network: the neuron in a Multi-Layer Perceptron (MLP). As the fundamental building block of neural networks, studying neurons provides insights into the foundational mechanisms of neural computation. Subsequently, we demonstrate how MSA can be applied to models with billions of parameters. Specifically, we examine the experts in a Mixture of Experts (MoE) model within a large language model (LLM) with 56 billion parameters. Finally, we demonstrate using MSA for multidimensional contributions by calculating pixel-wise contributions in a generative network. We also highlight complementary work \cite{fakhar2024downstream} showing how downstream processing dissociates neural activity from causal contributions in time series models.

By focusing on internal unit contributions across diverse tasks and models, MSA fills a key methodological gap in XAI. It provides a flexible toolset for probing neural computation beyond inputs, enabling us to not only explain, but also optimize these powerful models.

\section{Results}\label{results}
\subsection{Neural computations in a multi-layer perceptron }\label{ResMLP}
Our experimental setup involved a three-layer Multi-layer Perceptron (MLP)\cite{rumelhart_learning_1986} architecture trained on the MNIST dataset to classify digits\cite{lecun_gradient-based_1998}. We explored variations in the network by modifying the number of neurons and incorporating different regularization techniques\cite{ng_feature_2004}, achieving over 90\% accuracy across all configurations. Utilizing MSA, we quantified the contributions of each neuron not only to overall accuracy but also to digit-specific accuracy.

\begin{figure}[ht]
    \centering
    \includegraphics[width=\textwidth]{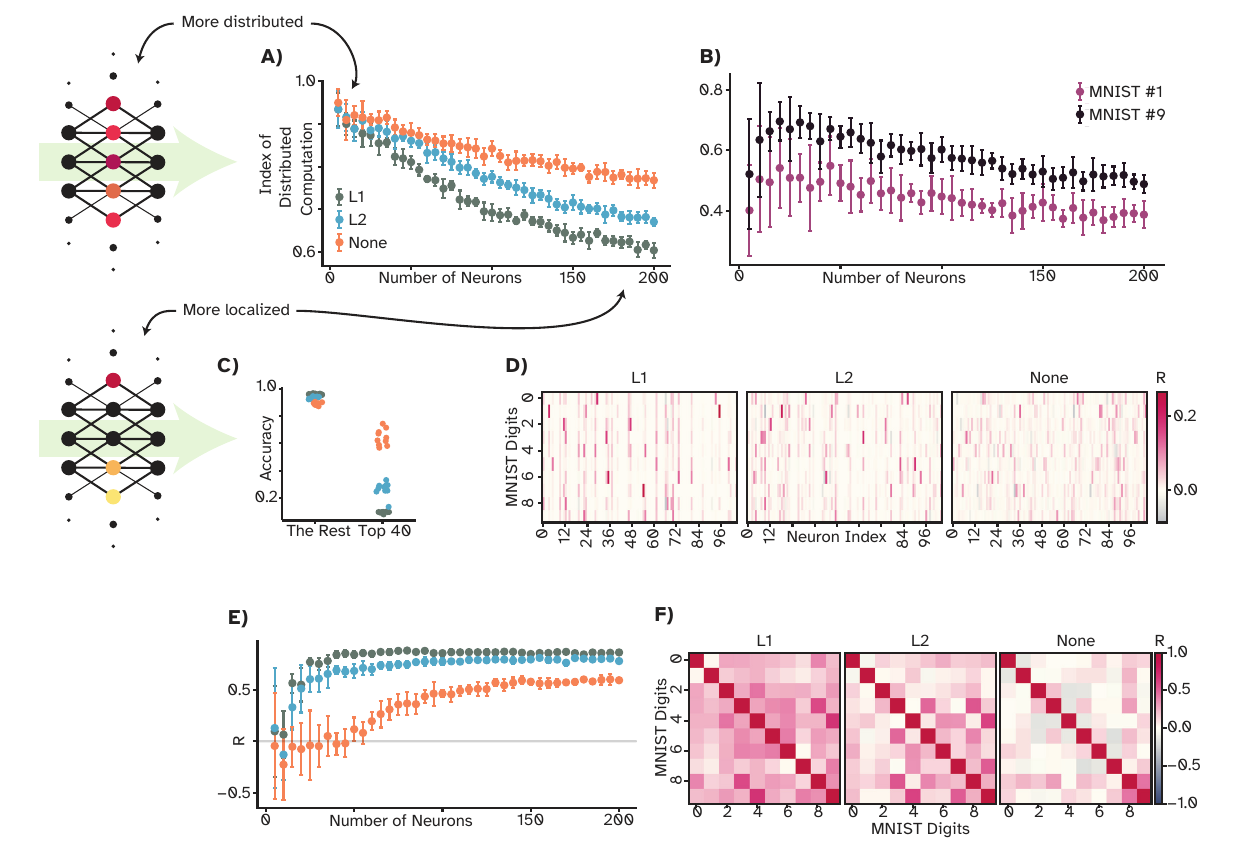}
    \caption[Results of Applying MSA on an MLP]{
        \textbf{A} Average index of distributed computation for MLPs with varying numbers of hidden neurons under different regularization schemes. L1 regularization results in few active neurons as neuron count increases, while no regularization keeps most neurons active. 
        \textbf{B} Average index of distributed computation for MLPs with varying numbers of hidden neurons for MNIST digits 1 and 9. More neurons are active for complex digits such as 9 compared to simpler ones like 1.
        \textbf{C} Impact on accuracy after removing the top 40 contributing neurons vs. the least contributing 160 neurons in a 200-neuron MLP under different regularization schemes.
        \textbf{D} Digit-specific contributions of each neuron in a 100-neuron MLP for three different regularization schemes, illustrating how regularization affects the distribution of digit recognition across neurons. 
        \textbf{E} Average correlations between absolute mean weights of neurons and their Shapley Values for MLPs with varying numbers of hidden neurons under different regularization schemes. Note the near-zero correlation for up to 50 neurons without regularization, consistently lower than L1 and L2 regularization. 
        \textbf{F} Correlation matrix among MNIST digits based on neuron contributions to digit classification. High correlations are observed between visually similar digits (e.g., 9 and 4, 8 and 3). 
    }
    \label{fig:mlp_figure}
\end{figure}

\subsubsection{Effects of regularization on the distribution of computation}
This experiment used an MLP with 200 neurons in the hidden layer. We trained the model under three regularization schemes: L1, L2, and no regularization, repeating each experiment for 10 trials. MSA was used to calculate how much each neuron in the hidden layer contributed to the overall and digit-wise accuracy measures.

Our findings revealed that most neurons did not contribute to accuracy in the presence of regularization (see Figure \ref{fig:mlp_figure}C). Removing the top 40 most contributing neurons from the network trained with regularization caused the accuracy to drop to chance level, with the effect being more pronounced for L1 regularization than L2. In contrast, removing the top 40 neurons from the unregularized network only reduced the accuracy to 50\%, indicating a less severe impact. Furthermore, removing the least-contributing 160 neurons did not affect the performance of the regularized models, while the unregularized model showed a slight decrease in performance. This observation highlights the robustness of unregularized models against internal failure due to a more distributed allocation of functions. These results also demonstrate that MSA effectively quantifies the contributions of neurons.

To quantify the distribution of computational processes within these networks, we introduced a metric termed ‘Index of Distributed Computation’ $(D)$, which is based on the concept of entropy from information theory (see section \ref{IDC}). $D$ spans from 0 to 1, with higher values indicating a more distributed computation across the network. Analysis in Figure \ref{fig:mlp_figure}A indicated that $D$ decreases with increasing number of neurons in the neural network. However, $D$ remained higher in non-regularized networks than their regularized counterparts, regardless of neuron count, suggesting a default tendency to distribute functions uniformly among the neurons. In contrast, both L1 and L2 regularized networks exhibited a greater decrease in $D$ with increasing network size, indicating a substantially constrained use of resources due to regularization. 

Furthermore, we observed the emergence of highly contributing multi-functional hubs in networks subjected to regularization, e.g., in a 100-neuron network in Figure \ref{fig:mlp_figure}D. In the absence of regularization, contributions were more evenly distributed among neurons, whereas L1 regularization led to the formation of hubs significantly involved in multiple tasks (evident as vertical strips), with other neurons contributing minimally. Interestingly, non-regularized networks displayed a more suboptimal allocation of function, with some neurons negatively impacting certain digits while aiding others, as evidenced by their negative contributions.

\subsubsection{Large weights do not always equal high importance}
A common assumption in neural network analysis is that the magnitude of a neuron's weight correlates with its importance \cite{frankle2018lottery, han2015learning}. We challenged this assumption by investigating the relationship between a neuron's importance, as measured by its contribution to the performance measured using MSA, and the mean of its absolute incoming and outgoing weights.

Our findings, illustrated in Figure \ref{fig:mlp_figure}E, show that when the MLP is trained without regularization, there is almost no correlation between weights and neuron importance if the number of neurons is just sufficient to perform the task optimally (around 50 neurons in this case). This finding indicates that in minimally sufficient networks, the contribution of neurons to the output is not dictated by their weight magnitude. As the number of neurons increases beyond the optimal count for task performance, the correlation between weight magnitude and importance rises. Interestingly, under regularization, especially L1, we observed a high correlation between neuron weights and importance, regardless of the number of neurons.

Our findings can be interpreted via multiple explanations involving network architecture, training dynamics, and regularization effects. One possibility is that in minimally sufficient, unregularized networks, every neuron contributes to performance, thus weight magnitude is a weak proxy for importance, because influence is distributed across many parameters. In over-parameterized models, functional redundancy emerges, allowing a subset of neurons to dominate. These neurons accumulate larger weights, the correlation between weight magnitude and functional importance increases, and the effective connectivity graph becomes sparser. L1 and L2 penalties accentuate this trend by driving small weights toward zero, isolating the most essential neurons irrespective of network size.

Training dynamics such as learning rates, initialization schemes, and data distribution could also influence how weights evolve, affecting the correlation between weight magnitude and neuron importance. Future work exploring these factors systematically may shed more light on the underlying mechanisms driving the relationships.

\subsubsection{Quantifying computational complexity and inter-class similarity}
MSA extends beyond assessing individual neuron contributions; it also quantifies the functional complexity of— and inter-class similarity between—the neural network's computations. To illustrate this aspect, we treated the classification of the ten MNIST digits as ten class-level functions and examined how the network allocates its computational resources across them.

We define inter-class similarity as the extent to which two class-level functions reuse the same neurons. Practically, we compute class-wise neuron contributions and correlate these vectors: higher overlap implies greater similarity. In a three-layer MLP with 50 hidden neurons, digits that are visually alike (e.g., 4 and 9) showed the strongest correlation in their neural contributions (Figure \ref{fig:mlp_figure}F).

We define functional complexity with the “Index of Distributed Computation” $D$, which reflects how many neurons must cooperate to solve a given class. Tasks engaging more neurons have higher $D$. When we compute $D$ for each MNIST digit across networks of varying size, digit 9 (complex curves) yielded the highest $D$, whereas digit 1 (a straight line) yielded the lowest (Figure \ref{fig:mlp_figure}B).

These analyses are not confined to simple datasets; the same approach scales to deep architectures and diverse domains. Neural models routinely solve many subtasks to accomplish a primary goal; for instance, a text-to-speech system must process dozens of phonemes. By measuring the complexity of each phoneme and the correlations between them, MSA could inform applications ranging from speech-therapy diagnostics to linguistic research and other healthcare uses \cite{peplinski_objective_2019, ravizza_relating_2001}.

\subsection{Language, arithmetic, and knowledge experts in an LLM}
We extended our analysis to Large Language Models to demonstrate the framework's scalability. Mixtral 8x7B is an open-source Large Language Model (LLM) based on the Transformer architecture \cite{jiang_mixtral_2024, vaswani2017attention}. It is proficient in five languages: French, German, Spanish, Italian, and English, and employs a sparse mixture of experts (SMoE) model\cite{jacobs1991adaptive, fedus2022switch, shazeer2017outrageously} with eight experts at each of its 32 layers, totaling 256 experts. The Mixture of Experts technique in AI involves a set of specialized experts, orchestrated by a routing network, to handle different parts of the input space, optimizing for both performance and efficiency.

We aimed to observe how the different experts in the model contribute to various domains. Three domains, arithmetic operations, language identification, and factual knowledge retrieval, were selected for analysis.

For the arithmetic domain, we had the model perform addition, subtraction, and multiplication of large numbers. In the language identification domain, we provided sentences in each of the five languages the model is proficient in and asked it to identify the language. For the factual knowledge retrieval domain, we asked the model to provide the capital cities of 20 given countries.

Using MSA, we calculated the contributions of each of the 256 experts to these tasks. No fine-tuning was performed on the model prior to this analysis. Indices mentioned in the following results follow zero indexing.

\begin{figure}[ht]
    \centering
    \includegraphics[width=\textwidth]{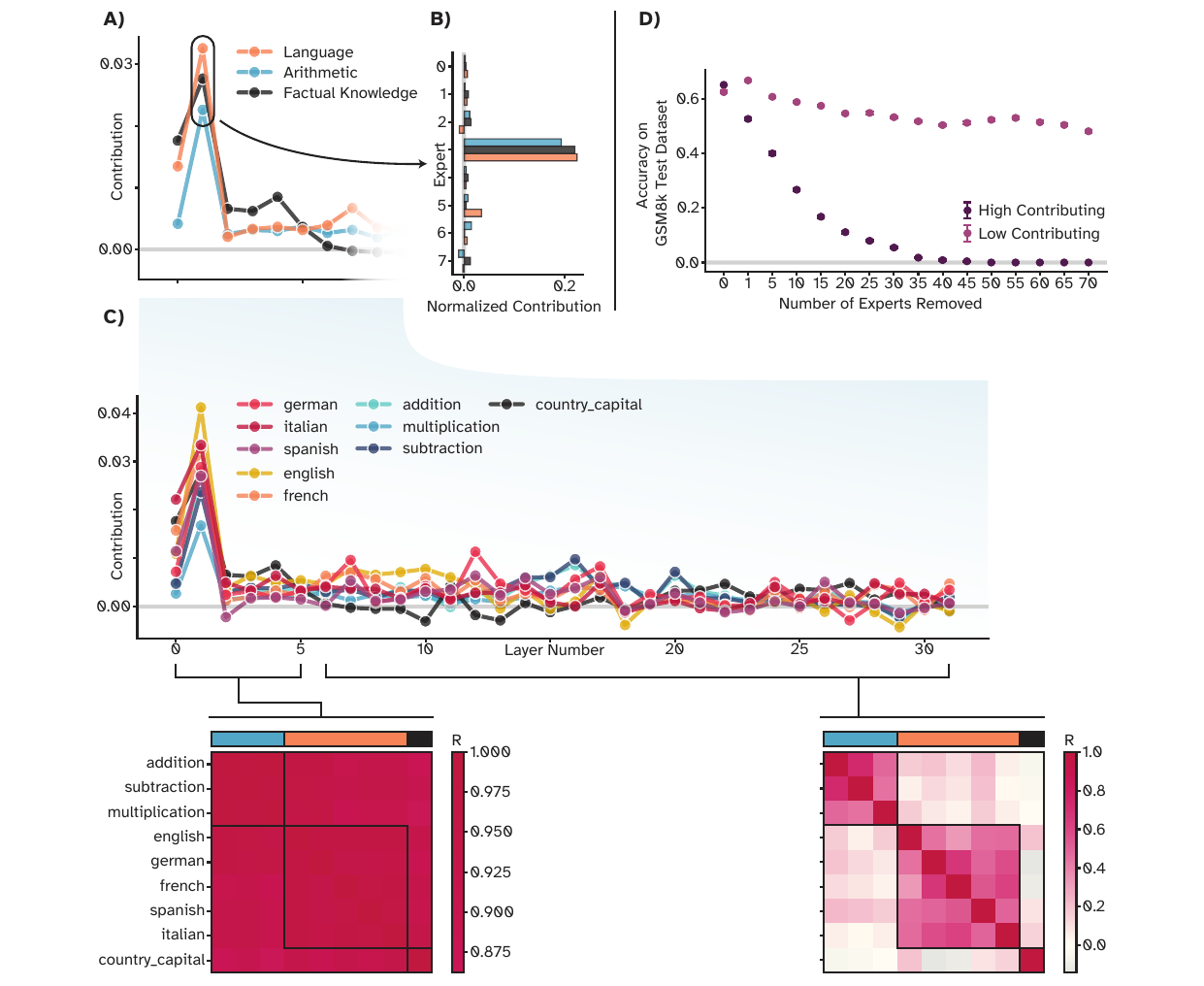}
    \caption[Results of Applying MSA on an LLM]{
    \textbf{A} Average contributions of experts across each layer within three distinct domains: Arithmetic, Language, and Factual Knowledge.  
    \textbf{B} Zoom-in on \textbf{A} to show the contributions of all eight experts in the first layer, emphasizing the dominant role of Expert 3 across all domains. The contributions of other experts in this layer are minimal in comparison.  
    \textbf{C} Further zoom-in to illustrate the average contributions of experts across layers for nine specific tasks, grouped into three domains: Arithmetic (addition, multiplication, subtraction), Language (French, German, Spanish, Italian, and English), and Factual Knowledge (country-capital identification). This panel highlights the differential importance of specific layers to each task. It also includes two correlation matrices showing the task-wise correlation of expert contributions in layers 0–5 and layers 6–31, respectively.  
    \textbf{D} Shows the LLM’s accuracy on the GSM8K dataset after progressively removing experts, comparing the effect of removing the least vs. the most contributing experts.
    }
    \label{fig:llm_figure}
\end{figure}


\subsubsection{Different layers are more involved for different domains}

Our analysis, shown in Figure \ref{fig:llm_figure}A and C, reveals that different layers in the model contribute differently depending on the task domain. By averaging the contributions of all experts in each layer, we found that the initial layers were consistently important across all domains, likely because they handle general-purpose processing shared by most tasks. However, beyond these early layers, we saw clear differences. The first and last layers were the most important for factual knowledge tasks, while the middle layers contributed very little. In contrast, arithmetic tasks relied heavily on the middle layers, suggesting that these layers are involved in the more complex computations required for mathematical reasoning. Language tasks followed yet another pattern, where the first few layers contributed the most, followed by a moderate contribution from the middle layers, reflecting a mix of low-level pattern recognition and higher-level processing.

To better understand how expert contributions vary across tasks, we also examined the task-wise correlations based on expert-wise contributions in two-layer groups: the early layers (0–5) and the later layers (6–31). In the early layers, we found that the expert contributions were highly correlated across all tasks, indicating that these experts are general-purpose and not specialized for any specific domain. However, in the later layers, while the correlations remained high within tasks of the same domain, they dropped sharply across different domains. This finding suggests that experts in the deeper layers become highly domain-specific, specializing in processing information relevant to a particular type of task.

\subsubsection{One very important expert}
We observed that the average contribution of the first layer is exceptionally high compared to all other layers, despite it not being the initial input layer. Further investigation in Figure \ref{fig:llm_figure}B revealed that the third expert in this first layer was highly contributing across all domains. Remarkably, the LLM's performance across these domains does not significantly drop even after removing all other experts in this layer. Whereas removing the third expert led to a 16\% drop in accuracy from 87.8\% average accuracy over all tasks. This phenomenon suggests the presence of redundant experts in the model. 

\subsubsection{Presence of redundant experts}
To investigate the potential presence of redundant experts, we extended our analysis beyond our original tasks to include more complex, real-world problems. We utilized the GSM8K dataset, comprising 8.5K high-quality, linguistically diverse grade school math word problems \cite{cobbe2021gsm8k}. This dataset allowed us to evaluate how removing both low and high-contributing experts (to the arithmetic tasks) affects the 3-shot accuracy of the LLM in solving these problems. The LLM performs with an accuracy of approximately 65\% on the GSM8k dataset.

Our findings, as illustrated in Figure \ref{fig:llm_figure}D, revealed some unexpected results regarding the role of experts in the model's performance on the GSM8K dataset. Notably, removing the lowest contributing expert increased the model's accuracy slightly above the original 65\%. This counterintuitive outcome suggests that this particular expert may have been negatively impacting performance, perhaps due to the routing network occasionally selecting it when it was not optimal. In the Mixture of Experts (MoE) architecture, the routing network is designed to assign tokens to the most appropriate expert. The fact that an expert can detract from accuracy indicates that the routing mechanism is not always making the best choices, potentially routing inputs to less suitable experts.

When we removed the five least contributing experts to arithmetic, the model's accuracy decreased by an average of 4\%. This drop implies that some low-contributing experts might hinder performance while others still support the network's overall function. Interestingly, even after removing 70 experts, which constitutes about 27\% of the total, the accuracy only declined by 14\%. This relatively modest decrease suggests that the model has a degree of resilience and can compensate for the loss of numerous lower-contributing experts, possibly due to redundancy among these experts or the network's ability to redistribute tasks among the remaining ones.

In contrast, removing just one high-contributing expert reduced accuracy comparable to that caused by removing the least-contributing 70 experts. This finding highlights the significant impact that top-performing experts have on the model's ability to solve complex problems. Moreover, when we removed the top 40 experts, the model's accuracy dropped to zero. This dramatic decline underscores that these experts are critical to the model's performance. The network fails to function effectively on the GSM8K dataset without them.

\subsection{Pixel-wise contributions in a generative network}
Finally, to demonstrate how MSA can be used for multidimensional network outputs, we trained a Deep Convolutional Generative Adversarial Network (DCGAN)\cite{goodfellow_generative_2014, radford_unsupervised_2016, dumoulin2016guide} on the CelebA\cite{liu2015deep} dataset to generate faces. The network consisted of six transposed convolutional layers with a total of 1987 filters. We chose a simple network for this proof-of-concept study rather than using a sophisticated model such as StyleGAN\cite{karras2019style}.

\begin{figure}[ht]
    \centering
    \includegraphics[width=\textwidth]{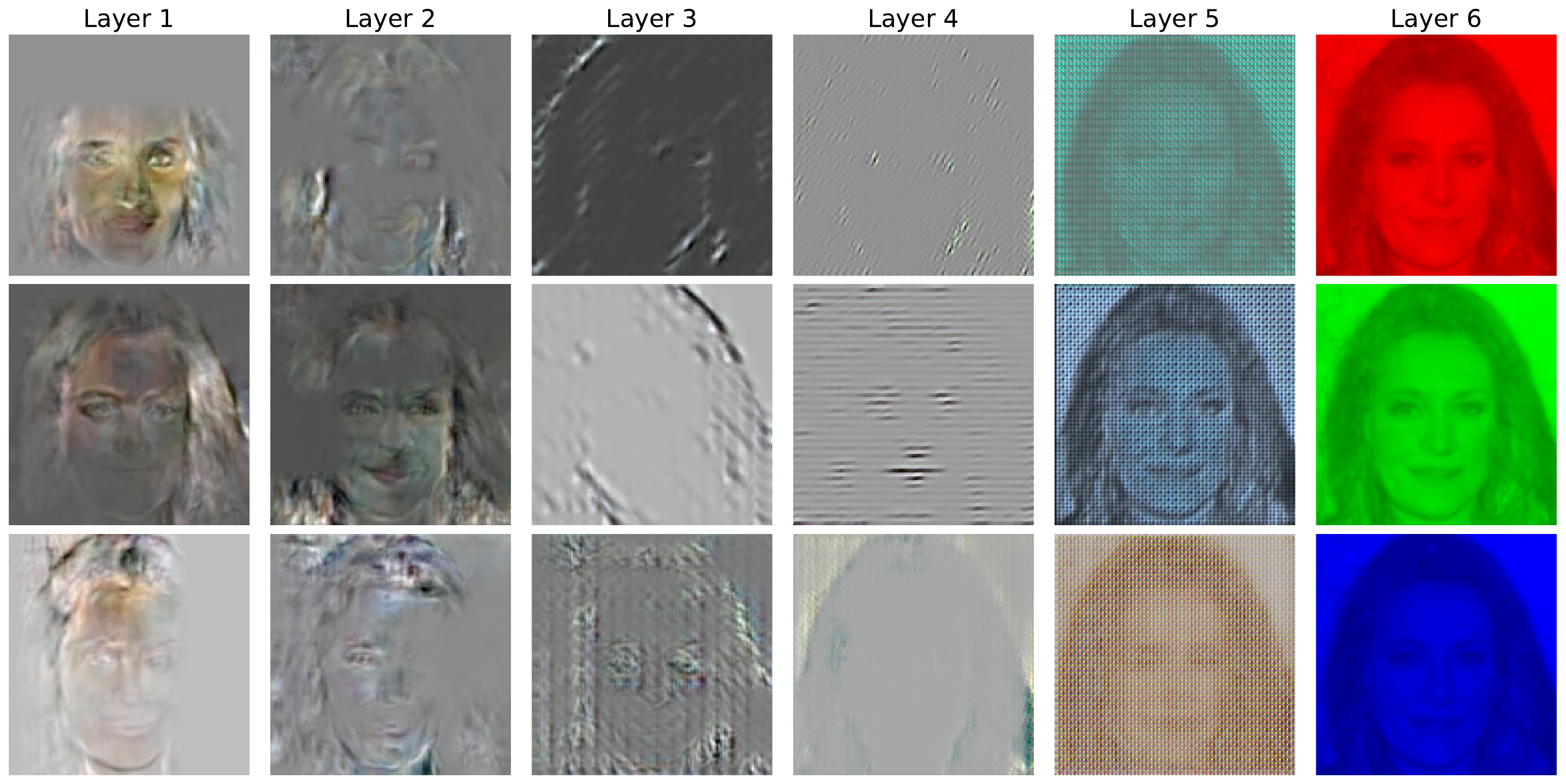}
    \caption[Contributions of three random neural units in each layer of a DCGAN]{Visualization of the contributions of three selected transposed convolutional filters from each layer in a DCGAN. The contributions have been normalized for better visualization. The first layer filters in the DCGAN contribute to higher-level facial features, while later layers focus on lower-level features like edges and textures, contrasting with CNNs used for classification, where early layers process low-level features and later layers handle higher-level features.}
    \label{fig:layerwise_contrib}
\end{figure}

MSA was applied to calculate the contributions of the transposed convolutional filters to a set of 32 randomly generated faces produced by the DCGAN. Figure \ref{fig:layerwise_contrib} showcases the contributions of three example filters from each layer, normalized for visualization purposes.

\subsection{Inversion of feature processing in GAN}

Interestingly, the filters in the first layer contribute to higher-level facial features, while those in the later layers contribute to lower-level features such as edges and textures. Specifically, the three filters in the last layer are dedicated to producing the Red, Green, and Blue colors, given that the network generates RGB images. This behavior contrasts with the filters in a Convolutional Neural Network (CNN) used for classification, where filters in the first layer typically process low-level features like edges and textures, and filters in the later layers process higher-level features\cite{zeiler_visualizing_2014, krizhevsky_imagenet_2012}.

The inversion of feature processing in generative networks could be attributed to the hierarchical nature of image synthesis. In the generative process, the network starts with an abstract representation of the desired image, which includes high-level structural features such as the overall shape of the face and major facial landmarks. As the image is gradually refined through successive layers, the network adds more detailed and specific features, such as textures and colors. Consequently, the early layers focus on broader, high-level features, while the later layers refine these structures with finer details.

\begin{figure}[ht]
    \centering
    \includegraphics[width=\textwidth]{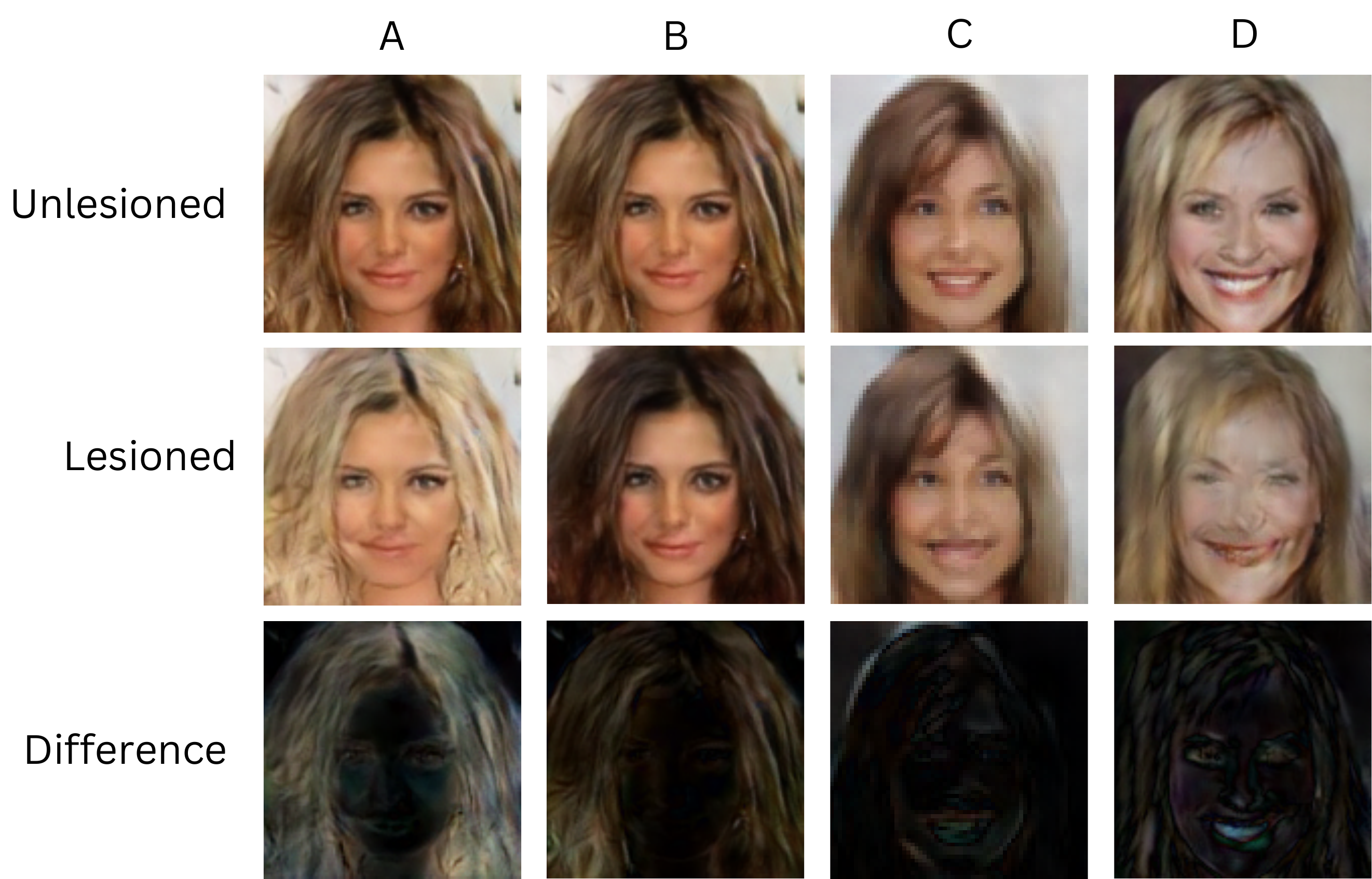}
    \caption[Clustering neural units based on their Contributions in a DCGAN]{The effect of removing clusters of transposed convolutional filters (lesioning) in a DCGAN. The first row displays images from an unlesioned DCGAN. The second row shows how the images change post-lesioning, while the third row highlights the differences between lesioned and unlesioned images. Column \textbf{A} demonstrates removing a cluster responsible for brown hair, resulting in the subject appearing blond. Column \textbf{B} shows how removing a cluster for blond hair results in brunette hair. Column \textbf{C} illustrates the removal of a cluster responsible for teeth, changing a smiling face with visible teeth to one smiling with closed lips, demonstrating the network's ability to maintain realism despite the modification. Column \textbf{D}  Removing clusters responsible for both eyes and teeth results in a closed-lip smile but creates an unrealistic dark shape around the eye region, illustrating the limits of the network's adaptability.}
    \label{fig:cluster_features}
\end{figure}

\subsection{Robustness in GAN}

We observed that different filters contributed to specific parts of the images; some were dedicated to generating eyes, others to teeth, etc. We observed targeted changes in the generated images by clustering these filters based on their contributions to specific image features and subsequently `lesioning' or removing these clusters.

As shown in Figure \ref{fig:cluster_features}A, removing a cluster responsible for brown hair resulted in the generator producing images of blonde individuals instead. Conversely, Figure \ref{fig:cluster_features}B demonstrates that removing a cluster associated with blonde hair led to the generation of brunette subjects. These observations suggest the network has learned to separate hair colors into distinct filter clusters.

Interestingly, when we removed a cluster of filters responsible for generating teeth, the network adapted by producing faces with closed-lip smiles instead of open-mouthed smiles showing teeth (Figure \ref{fig:cluster_features}C). Despite this significant modification, the generated images maintained a realistic appearance. This adaptability can be attributed to the adversarial training process of the GAN \cite{goodfellow_generative_2014}. Even with part of the network responsible for teeth generation removed, the generator still produced images that could potentially deceive the discriminator network, highlighting the robustness against lost neural units.

However, the network's ability to compensate for lost neural units had limits. As illustrated in Figure \ref{fig:cluster_features}D, when we removed clusters responsible for both eyes and teeth, the generator produced images with closed-lip smiles but struggled with the eye region, creating a strange, dark shape around where the eyes should be.

A possible explanation could be that hair color, being a relatively simple feature, can be easily substituted when its corresponding filters are removed. Smiles, while more complex, still have a natural alternative (closed-lip smiles) that the network can fall back on when teeth-generating filters are removed. Eyes, however, are a crucial and complex facial feature without a simple alternative. When eye-generating filters are removed, the network lacks the necessary components to create a realistic substitute. Instead, it attempts to fill the space with a dark patch, likely representing the best approximation it can manage given the available filters.

\section{Discussion}\label{discussion}
The present paper introduces a general-purpose framework to quantify neural contributions in any neural network, known as Multiperturbation Shapley-Value Analysis (MSA). We have released this framework as an open-source Python package. Through three case studies, including a highly complex LLM with 56 billion parameters, we demonstrated the utility of MSA. Additionally, we showed how MSA can be used to quantify the complexities and similarities among tasks that a neural network is trained on, highlighting the practical applications of this information.

However, MSA can be computationally expensive for large models, since it systematically removes neural units and evaluates their impact on the model's output. For a model as large as Mixtral 8x7B, with 56 billion parameters, it took just under nine days to run on a single A100 80GB Nvidia GPU. A future enhancement for our package is the integration of permutation sampling strategies, as introduced by \cite{mitchell2022sampling}, to accelerate computation.

 In neural networks, the function of one neuron may depend on another, and two neurons might have overlapping functions (redundancy). This interdependency cannot be discerned from their individual contributions alone. Therefore, MSA can be used to analyze interactions between neuron pairs  \cite{keinan_fair_2004}. Although not discussed in this paper, we plan to demonstrate and extend this functionality to multiple dimensions, similar to our approach with Shapley Modes, in future work.


It is important to note that while MSA provides insights into \textit{what} each unit does, it does not elucidate \textit{how} it does it. For example, in the Mixture of Experts model, MSA can identify each expert's contribution to specific tasks, but does not offer information on the underlying parameters or architectural elements that enable the expert to perform those tasks. Understanding the mechanisms behind these contributions remains an open area for further research, which could lead to deeper insights into the functioning of neural networks and potentially guide the design of more efficient and interpretable models.

\section{Methods}\label{Methods}
\subsection{Multi-perturbation Shapley-Value Analysis}
Multi-perturbation Shapley-Value Analysis (MSA) is a game-theoretical framework based on the concept of Shapley Values, introduced by Lloyd Shapley in the 1950s \cite{shapley1953value}. Shapley Values provide a method for fairly distributing the total gains (or costs) among players in a cooperative game based on their individual contributions.

In simpler terms, the Shapley Value of a player represents how much that player contributes to the overall outcome of the game. It is calculated by considering all possible coalitions (subsets of players) and averaging the marginal contributions of the player across these coalitions. The Shapley Value ensures that each player is allocated a contribution that reflects their impact on the game's outcome.

The Shapley Value is defined by a set of axioms that make it a particularly useful tool in game theory and cooperative scenarios. Its foundational axioms ensure a fair distribution of the total value among the players, encompassing several key properties:
\begin{enumerate}
    \item \textbf{Efficiency:} The sum of the Shapley Values for all players equals the total value obtained by the grand coalition of all players.
    \item \textbf{Symmetry:} Players who contribute equally to every possible coalition receive identical Shapley Values.
    \item \textbf{Dummy Player:} Any player who does not contribute additional value to any coalition receives a Shapley Value of zero.
    \item \textbf{Additivity:} For any two games, the Shapley Value of their sum is the sum of their Shapley Values.
\end{enumerate}

The Shapley Value is unique in that it is the only method that satisfies all four of these properties simultaneously. If a value distribution method fulfills these four axioms, it must be the Shapley Value.

In the context of this paper, the `game' is any trained neural network model $\mathcal{M}$ with $n$ neural units: $N = \{m_1, m_2, ..., m_n\}$ as players. The value $V(S)$ of the `game' represents the performance of the neural network with $S \subseteq N$ neural units.  Traditionally, Shapley Analysis uses performance metrics like accuracy, mean square error, etc., to calculate contributions.

We extend this concept to accommodate multiple performance metrics or a multidimensional output of the neural network. For the sake of explanation, let's assume a generator neural network that generates faces. In this generator model, the players are the neural units, and the output $V(S)$ is the 32 RGB images of the faces generated by the generator network. The output $V(S)$ is a tensor of shape $(32, 3, W, H)$ where $W$ and $H$  represent the width and height of the images, respectively.

In the context of multidimensional outputs, we introduce the term Shapley Modes to describe the contributions of each neural unit. In the same generator network example, the Shapley Mode $\gamma(m_i)$ of the neural unit $m_i$ is also a tensor of the same shape $(32, 3, W, H)$ as the output. This tensor illustrates how the neural unit contributes to each pixel of the 32 faces generated by the DCGAN.

To define the Shapley Mode mathematically, we first define the marginal contribution of a neural unit \( m_i \) to a coalition \( S \), where \( m_i \notin S \), as:
\begin{equation}
\Delta_i(S) = V(S \cup \{m_i\}) - V(S),
\label{eq:marginal_contribution}
\end{equation}
where \( V(S) \) represents the value function for the coalition \( S \).

The Shapley Mode \( \gamma(m_i) \) is then defined as:
\begin{equation}
\gamma(m_i) = \frac{1}{n!} \sum_{R \in \mathcal{R}} \Delta_i(S_i(R)),
\label{eq:shapley_mode}
\end{equation}
where \( \mathcal{R} \) denotes the set of all possible permutations of the players, and \( S_i(R) \) represents the set of players preceding \( m_i \) in a given permutation \( R \). This formulation ensures a fair and comprehensive evaluation of the contribution of each neural unit across all possible coalitions.

where $\mathcal{R}$ is the set of all $n!$ orderings of $N$ and $S_i(R)$ is the set of players preceding $m_i$ in the ordering $R$. For multidimensional outputs $V(S)$, all mathematical operations like addition and subtraction are done element-wise. Essentially, MSA works by systematically removing neural units (perturbation) and capturing their causal contributions to the overall outcome.

\begin{figure}[ht]
    \centering
    \includegraphics[width=\textwidth]{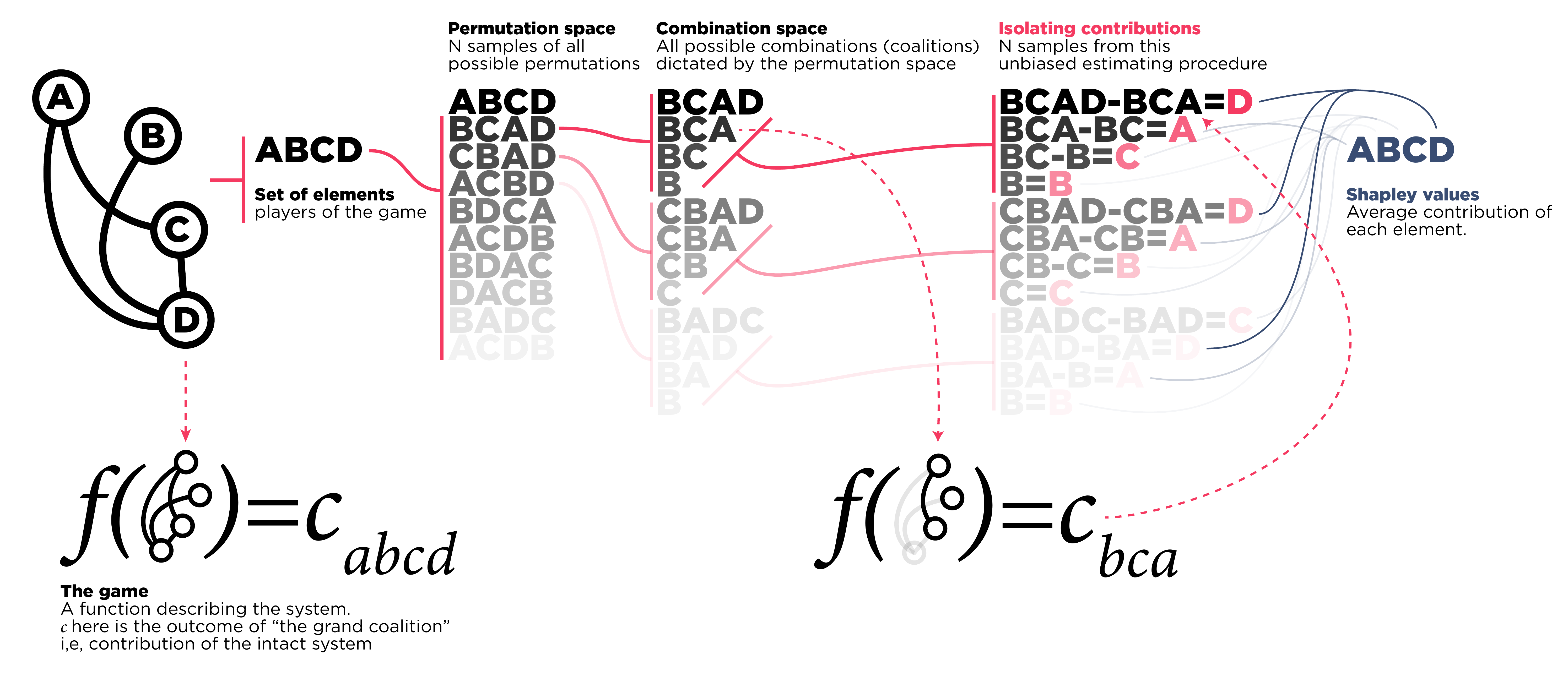}
    \caption[Graphical Illustration of the working of Multi-Perturbation Shapley-Value Analysis]{
    This is a graphical illustration of the MSA algorithm. MSA samples the space of all possible combinations of neural unit groupings to estimate their causal contributions (red). To do so, it first permutes the players and expands each permutation configuration to dictate which combinations should be perturbed. In this schematic example, to acquire the outcome produced by BCA, the example player D was removed from the game. Then to have the outcome of BC, nodes D and A were perturbed. Therefore, MSA produces a multi-site perturbation dataset that contains the outcome produced by potentially tens of thousands of unique neural unit groupings. To then isolate the contribution of individual neural units to each of these groupings, MSA contrasts two cases where the target node was perturbed and where it was not. The causal contribution of each neural unit to the network's outcome is then the average of the neural unit's contributions to all groupings.
        }
    \label{fig:msa_algo}
\end{figure}

Calculating Shapley Modes by considering all possible orderings is impractical due to the factorial growth of possible orderings. Therefore, we use the Monte Carlo estimation to calculate Shapley Modes. By sampling $p$ orderings from the set of all $n!$ orderings, we can obtain an unbiased estimation of the Shapley Modes. Figure \ref{fig:msa_algo} describes this process in detail.

\subsection{MSA applied to Neural Networks}
This section discusses how we applied MSA to the neural networks examined in this paper. We detail the perturbation methods and technical aspects of analyzing Shapley Modes.

\subsubsection{MSA applied to MLP}
The Multi-layer Perceptron (MLP)\cite{rumelhart_learning_1986} is one of the most fundamental neural network architectures, commonly used as a starting point for understanding neural computation and learning dynamics in AI. In this study, we used an MLP with a three-layer architecture—comprising an input layer, a hidden layer, and an output layer—to understand the computations performed by the neurons under various configurations. Specifically, we manipulated the number of neurons in the hidden layer, ranging from a minimal configuration of 5 to a more complex setup with 200 neurons. This variance in neuron count helps in understanding the scalability and the dynamics of neural contribution across different network sizes.
 
Each MLP was trained under three different regularization schemes\cite{ng_feature_2004}:
\begin{itemize}
    \item \textbf{L1 Regularization:} This method adds a penalty equal to the absolute value of the magnitude of coefficients. L1 regularization can lead to a sparse model where some weights become almost zero.
    \item \textbf{L2 Regularization:} Also known as Ridge Regression, L2 adds a penalty equal to the square of the magnitude of coefficients. This method primarily helps in preventing overfitting by ensuring the weights do not grow too large.
    \item \textbf{No Regularization:} Training without any regularization provides a baseline to understand the network's behavior in its most natural form, without any constraints imposed on the weight values.
\end{itemize}

To ensure that our results were consistent and not due to random variations in network initialization or training data shuffling, we repeated the training process for 10 trials for each configuration. These MLPs were trained on the MNIST dataset, which is a large database of handwritten digits commonly used for training various image processing systems \cite{lecun_gradient-based_1998}. The MNIST dataset contains 70,000 images of handwritten digits (0-9), each a 28x28 pixel grayscale image.

We used MSA to calculate how each neuron in the hidden layer contributes to the accuracy for each of the 10 MNIST Digits. The perturbation was performed by setting the activation of the targeted neuron to 0.

\subsubsection{Index of Distributed Computations}\label{IDC}
To quantify the distribution of computation among neurons, we propose an entropy-based metric. Let \( c_i \) denote the contribution of the \( i \)-th neuron, which can be positive or negative. To reflect the magnitude of contributions without canceling out opposite signs, we define the squared contributions as \( s_i = c_i^2 \). The normalized contribution \( p_i \) is then computed as:
\begin{equation}
p_i = \frac{s_i}{\sum_{j=1}^{N} s_j},
\label{eq:normalized_contribution}
\end{equation}
where \( N \) is the number of neurons. The set \( \{p_i\} \) forms a valid probability distribution, with \( \sum_{i=1}^{N} p_i = 1 \).

The entropy \( H \), which measures uncertainty or diversity in a probability distribution, is defined as:
\begin{equation}
H = -\sum_{i=1}^{N} p_i \log p_i,
\label{eq:entropy}
\end{equation}
following the formulation by Shannon \cite{shannon1948mathematical}. A higher entropy indicates a more uniform distribution, while a lower entropy reflects dominance by a few components. To make the metric independent of the number of neurons, we normalize \( H \) with its maximum value \( H_{\text{max}} = \log N \), resulting in:
\begin{equation}
M = \frac{H}{H_{\text{max}}}.
\label{eq:normalized_entropy}
\end{equation}

A higher \( M \) implies a more distributed computation among neurons, while a lower \( M \) indicates that only a few neurons contribute significantly. The entropy \( H \) satisfies several desirable properties. First, it is invariant to permutations of \( \{p_i\} \), ensuring fairness irrespective of the order of neurons. Second, it is maximized for a uniform distribution, aligning with our goal of measuring how evenly computation is distributed. These properties make the entropy-based metric an effective tool for capturing the extent of computational distribution.

\subsubsection{MSA applied to LLM}
Language modeling involves predicting the next word or token in a sequence given the preceding words or tokens. Large Language Models (LLMs), like those based on the Transformer architecture \cite{vaswani2017attention}, work with tokens, which are units of text that can be as small as a single character or as large as a word or phrase. During training, the model learns to encode these tokens into high-dimensional vectors, capturing their contextual relationships. When generating text, the model takes a sequence of tokens as input and produces a probability distribution over the possible next tokens. The token with the highest probability is then selected as the next token in the sequence, and this process continues iteratively to generate coherent and contextually relevant text.

One approach to enhance the performance and efficiency of LLMs is the Mixture of Experts (MoE) technique \cite{jacobs1991adaptive, fedus2022switch, shazeer2017outrageously}. MoE models consist of multiple specialized sub-models, known as experts, each trained to handle different parts of the input space. A gating mechanism dynamically selects the most relevant experts for each input, allowing the model to leverage the strengths of various experts while reducing the computational load by not activating all experts simultaneously.

For this analysis, we used the Mixtral 8x7B model \cite{jiang_mixtral_2024}, an MoE model with 32 layers and eight experts at each layer, totaling 256 experts. The model uses only 2 out of 8 experts at each layer for each token. The model is also proficient in five languages: Italian, French, Spanish, German, and English. For our analysis, we used the pre-trained `instruct' version of the model \cite{ouyang2022training}. The `instruct' version of the model has been fine-tuned to follow prompted instructions. To expedite the MSA process, we quantized the model to 4 bits \cite{hubara2018quantized}. Quantization reduces the precision of the model's parameters, significantly speeding up computation while maintaining an acceptable level of accuracy.

MSA was used to calculate the contributions of the experts to the model's performance in three domains: arithmetic, language, and factual knowledge.

For the arithmetic task, we evaluated the model on three subtasks: addition, subtraction, and multiplication. We evaluated the model on 30 pairs of 7-digit whole numbers for the addition task. Similarly, we used 30 pairs of 7-digit whole numbers for the subtraction task. For the multiplication task, we evaluated the model on 30 pairs of 4-digit whole numbers because the product of these 4-digit numbers can be up to 8 digits.

The performance was assessed by calculating the overlap of digits between the model's prediction and the correct answer. Specifically, we used the Levenshtein distance \cite{1965LevDist}, which measures the minimum number of single-character edits (insertions, deletions, or substitutions) required to change one word into another. Levenshtein distance was chosen over metrics like mean squared error (MSE) because it does not account for the positional significance of digits. For example, if the model is off by one digit, the MSE error is much larger if the incorrect digit is in the millionth place compared to the one's place.

The instructions for all of these tasks were provided in English. The accuracy for the addition and subtraction tasks was around 90\%, indicating the model's strong performance in these areas. However, the accuracy for the multiplication task was lower, around 55\%, suggesting that multiplication of large numbers is a more challenging task for the model.

To find language-specific experts in the model, we evaluated the model on a language identification task. Fifteen sentences from each of the five languages the model is proficient in were provided to the model to identify the language. The instructions for these tasks were also given in the same language as the sentence provided. The sentences were sourced from the Tatoeba project \cite{Tatoeba}, which is a large database of sentences and translations in many languages. The model achieved 100\% accuracy for all languages except Spanish, where it incorrectly identified the language as Latin in 4 out of 15 instances. 

For the factual knowledge retrieval task, we evaluated the model by providing the names of 30 countries and asking for their capital cities. Unlike the other tasks, this was done in a one-shot manner. We provided Madrid, the capital of Spain, as an example to the model because it was not very accurate without an initial prompt. With this example, the model's accuracy was 70\%. The instructions for this task were provided in English.

We used MSA to calculate the experts' contributions toward these nine subtasks: Addition, Subtraction, Multiplication, French, English, German, Spanish, and Italian language identification, and Factual Knowledge Retrieval. The perturbation was performed by excluding the contribution of the targeted expert during computation. To estimate the Shapley Values, we sampled 1000 orderings from the total possible $256!$ orderings.

\subsubsection{MSA applied to DCGAN}
Generative Adversarial Networks (GANs)\cite{goodfellow_generative_2014} represent a sophisticated class of algorithms within unsupervised machine learning, featuring two neural networks—the generator and the discriminator—engaged in a zero-sum game. While the generator aims to produce data indistinguishable from genuine data, the discriminator strives to discern between real data and the generator's fabrications, thus continuously improving the fidelity of the generated outputs.

Deep Convolutional Generative Adversarial Networks (DCGANs)\cite{radford_unsupervised_2016} are a variant of the standard GAN architecture, primarily using convolutional and convolutional-transpose layers in the generator and discriminator, respectively \cite{dumoulin2016guide}. Transposed convolutional layers work by swapping the forward and backward passes of a convolution. In the context of this paper, the generator network is trained on the CelebA dataset \cite{liu2015deep} to generate random RGB images of faces.

The process starts with a random, low-dimensional input (like a noise vector), which acts as a seed for generating the image. This input passes through several layers of transposed convolutions, where each layer uses filters to progressively upscale the input and add finer details. By the time this input has passed through all the layers, it has transformed from a simple random noise vector into a complex, high-resolution image resembling a human face.

Each transposed convolutional filter in the generator refines certain aspects of the image, such as textures and contours, effectively learning from the dataset how to produce realistic faces from random noise, mimicking the variations seen in real human features. To understand the contribution of each filter, they were systematically perturbed as per MSA explained in the previous section. The perturbation was applied by setting the output of that filter to zeros.

MSA was used to calculate the contribution of each transposed convolutional filter toward each pixel of the 32 RGB Images generated by the generator network. The perturbation was performed by setting the output of the targeted  filter to zeros. The contributions for each filter calculated using MSA are of the same shape as the output of the generator network. Since the contribution is calculated for 32 randomly generated faces, the shape of the contributions is (32, 3, Height, Width), indicating how that filter contributes towards each pixel for those 32 faces.

For clustering these contributions, traditional clustering methods like k-means \cite{macqueen1967some, lloyd1982least}, t-SNE\cite{van2008visualizing}, and DBSCAN\cite{ester1996density} proved ineffective. This could be due to the high dimensionality of the data, the non-linear nature of the contributions, or the potential sparsity of the contribution patterns. To overcome these challenges, we used a different approach. Our approach involved calculating a similarity matrix using Pearson’s correlation \cite{pearson1895vii, lee1988thirteen} between all the filters based on their contributions. The absolute value of this matrix was then considered as an adjacency matrix of a graph. To cluster these filters into different groups, we applied a community detection algorithm called the Louvain Algorithm \cite{blondel2008fast}.

The Louvain Algorithm is a community detection method that optimizes modularity across the graph, effectively identifying densely connected nodes (in this case, filters with similar contribution patterns) and grouping them into distinct clusters. This method allowed us to discern which groups of filters were responsible for generating specific features in the faces, such as eyes, hair, or skin texture.

\section{Data and Code Availability}\label{datacode}
The Multiperturbation Shapley-Value Analysis (MSA) is implemented as an open-source Python package named msapy. The package can be accessed from the following GitHub repository: \url{https://github.com/kuffmode/msa}.

Additionally, the code for the three examples demonstrating the application of the MSA package for XAI in neural networks is available at: \url{https://github.com/ShreyDixit/MSA-XAI}. This repository includes all necessary scripts and instructions to replicate the examples discussed in this paper.

All data used for training the models in these examples is open-access. The repository also contains the code required to download and use the data for training purposes, ensuring that the experiments can be easily reproduced and verified.

\section{Acknowledgements}
The funding is gratefully acknowledged: SD: SFB 936-178316478-A1. KF: German Research Foundation (DFG)-SFB SPP 2041/GO 2888/2-2 and the Templeton World Charity Foundation, Inc. (funder DOI 501100011730) under grant TWCF-2022-30510. FH: DFG TRR169-A2. PM: N/A. KPK: NIH "Quantifying causality for neuroscience" 1R01EB028162-01. CCH: SFB 936-178316478-A1; TRR169-A2; SFB 1461/A4; SPP 2041/HI 1286/7-1, the Human Brain Project, EU (SGA2, SGA3)



\bibliography{sn-bibliography}

\end{document}